\pdfoutput=1

\documentclass[11pt]{article}

\usepackage{acl}

\usepackage{times}
\usepackage{latexsym}

\usepackage[T1]{fontenc}

\usepackage[utf8]{inputenc}

\usepackage{microtype}

\usepackage{times}
\usepackage{latexsym}
\usepackage{enumitem}
\usepackage{amssymb}
\usepackage{amsmath}
\usepackage{bbm}
\usepackage{multirow}
\usepackage{tabularx}
\usepackage{booktabs}
\usepackage{graphicx}
\graphicspath{{./img/}}

\title{Breaking Down Word Semantics from Pre-trained Language Models through Layer-wise Dimension Selection}

\author{Nayoung Choi \\
  \texttt{skdudenql11@naver.com} \\}

\begin{document}
\maketitle

\begin{abstract}
Contextual word embeddings obtained from pre-trained language model (PLM) have proven effective for various natural language processing tasks at the word level. However, interpreting the hidden aspects within embeddings, such as syntax and semantics, remains challenging. Disentangled representation learning has emerged as a promising approach, which separates specific aspects into distinct embeddings. Furthermore, different linguistic knowledge is believed to be stored in different layers of PLM. This paper aims to disentangle semantic sense from BERT by applying a binary mask to middle outputs across the layers, without updating pre-trained parameters. The disentangled embeddings are evaluated through binary classification to determine if the target word in two different sentences has the same meaning. Experiments with cased BERT$_{\texttt{base}}$ show that leveraging layer-wise information is effective and disentangling semantic sense further improve performance.
\end{abstract}

\section{Introduction}
Pre-trained language models (PLMs) like BERT \cite{devlin-etal-2019-bert} have been successful in generating contextual word embeddings for diverse word-level NLP tasks, such as named entity recognition, part-of-speech (PoS) tagging, coreference resolution, and word sense disambiguation (WSD) \cite{raganato-etal-2017-word}. The power of contextual word embeddings lies in their capacity to capture the overall traits of a word, enabling them to perform well across wide-ranging NLP tasks.

Although contextual word embeddings are capable of capturing diverse hidden aspects such as semantic sense, syntactic role, and sentiment, interpreting these factors remains a challenge. To address this issue, disentangled representation learning (DRL) (\citealp{do2019theory}; \citealp{9947342}) has emerged as an approach to separate specific aspects into distinct representations. DRL has shown promise in computer vision through methods such as InfoGAN \cite{chen2016infogan}, but its potential for NLP tasks has yet to be fully explored \cite{vishnubhotla2021evaluation}. In the text domain, \citet{xu2020variational} proposed a constraint on the latent space of variational autoencoder (VAE) with PLM to control sentiment and topic. \citet{DBLP:journals/corr/abs-2105-02685} utilized mutual information (MI) to disentangle aspects from text for fair classification and textual style transfer. The groundwork for this paper was laid by \citet{Zhang2021}, who trained binary masks to find subnetworks within BERT using masking transformers \cite{Zhao2020} to disentangle sentiment from genre, toxicity from dialect, and syntax from semantic. 

\begin{figure}[!]
    {\small
    \rule{7.7cm}{0.3mm} \\
    \noindent $w$ \texttt{= \textcolor{red}{raise} \\
    ${s_1}$ = We visited a farm where they \textcolor{red}{raise} chickens. \\
    ${s_2}$ = A few important questions were \textcolor{red}{raise}d after the attack.
    } \vspace{-0.1cm} \\ 
    \rule{7.7cm}{0.3mm}
    \vspace{-0.5cm}
    \caption{A sample labeled as false from the WiC dataset, indicating that the target word $w$ in sentences ${s_1}$ and ${s_2}$ do not correspond to the same meaning.}
    \label{fig:WiC example}
    \vspace{-0.7cm}
    }
\end{figure}

On the other hand, several studies have shown that linguistic knowledge like syntactic and semantic aspects are encoded in different layers of PLM (\citealp{jawahar-etal-2019-bert}; \citealp{tenney-etal-2019-bert}; \citealp{hewitt-manning-2019-structural}; \citealp{bommasani-etal-2020-interpreting}). Furthermore, each self-attention head seems to focus on different aspect of language information (\citealp{vig-belinkov-2019-analyzing}; \citealp{clark2019does}; \citealp{zhao-bethard-2020-berts}). However, it is common practice to represent a word simply as the sum of the last four layers' hidden states even for embedding-based WSD (\citealp{scarlini-etal-2020-contexts}; \citealp{SensEmBERT}; \citealp{loureiro2021analysis}). Despite its practicality and widespread usage, it overlooks the heterogeneous information captured by different layers. 

\begin{figure*}[htb!]
\centering
\includegraphics[width=\textwidth]{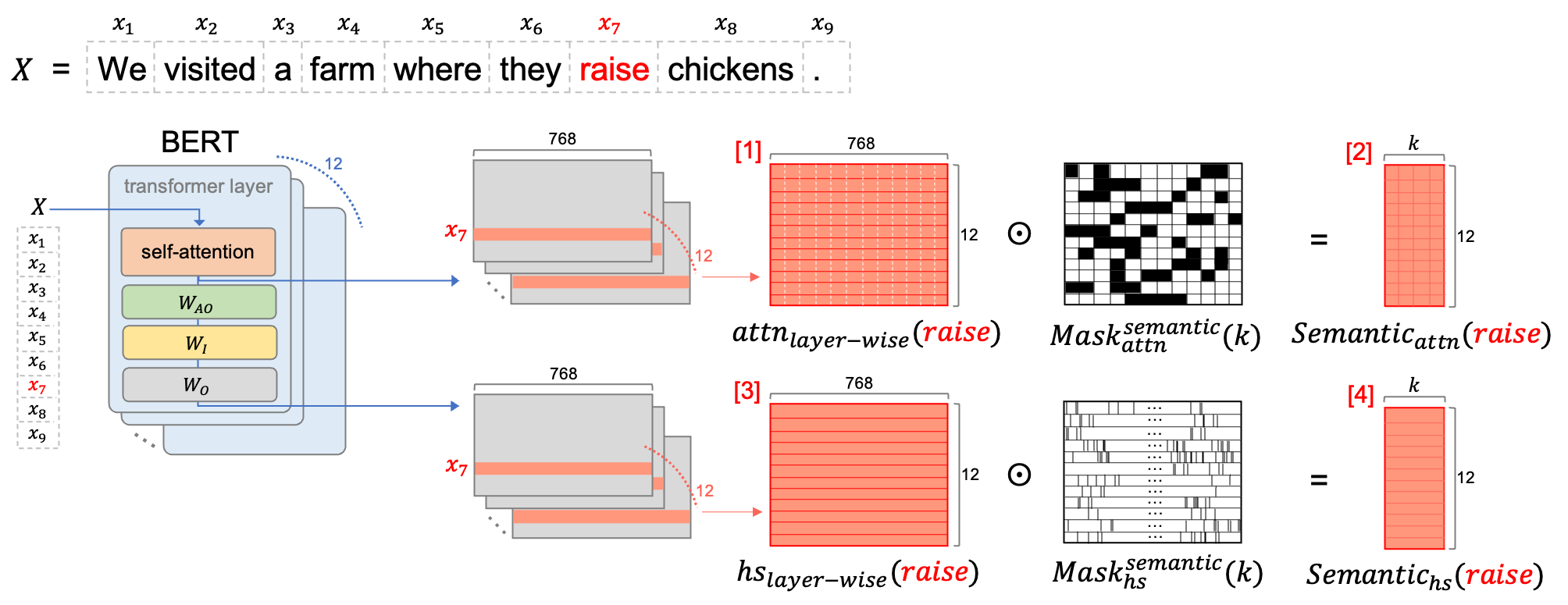}
\vspace{-0.5cm}
\caption{Taking BERT$_{\texttt{base}}$, it illustrates the way to disentangle semantic sense of the word $\textcolor{red}{raise}$ from middle outputs (layer-wise hidden states or self-attention outputs) of input sentence $X$. Binary trained $Mask_{*}^{semantic}(k)$ disentangles sense and reduces the output dimensionality from 768 to $k$ for each layer. $attn_{layer-wise}(x)$ is divided into 12 attention heads and $Mask_{attn}^{semantic}(k)$ masks it head-wise, while $Mask_{hs}^{semantic}(k)$ masks $hs_{layer-wise}(x)$ dimension-wise. If a word consists of more than one token, mean of tokens' outputs is used to represent a word.}
\label{fig:masker}
\vspace{-0.5cm}
\end{figure*}

This paper studies to disentangle the semantic sense of word by masking the middle outputs from BERT in a layer-wise manner. The disentangled embeddings are evaluated by binary classification task, specifically utilizing the WiC format (shown in Figure \ref{fig:WiC example}). The task has practical implications in real-world scenarios, such as re-ranking search engine results with relevance for ambiguous queries. The contributions of this work are as follows:
\begin{itemize}[topsep=0pt, partopsep=0pt]
\setlength\itemsep{-0.3em}
    \item The study presents a method to disentangle semantic sense from BERT, masking embedding dimensions without updating pre-trained model parameters.
    \item Layer-wise information is successfully used to determine the semantic similarity between the target word in two different sentences.
    \item It includes a comprehensive analysis about the behavior of the trained mask across the layers.
\end{itemize}

\section{Approach}
The aim is to disentangle semantic sense from layer-wise middle outputs about word obtained from pre-trained BERT (specifically, Cased BERT$_{\texttt{base}}$ is used). A binary mask is trained to select $k$ dimensions of hidden states or self-attention outputs (not the attention weights) for each layer, as illustrated in Figure \ref{fig:masker}. Using the subsets of them, the semantic similarities between the target word $w$ in sentences $s_1$ and $s_2$ are computed. Subsequently, for evaluation, a binary classifier is trained as a predictor-based metric \cite{Carbonneau} to determine whether the two $w$ share the same meaning or not. 

\subsection{Disentanglement by Masking}
\label{Disentanglement by Masking}
Considering a transformer layer of BERT \cite{Zhao2020}, the input $\mathbf{X}\in\mathbb{R}^{N \times h}$ ($N$: maximum token length, $h$: hidden size) undergoes multi-head self-attention. The outputs are then transformed by $W_{AO}$ and further passed to the next layer through $W_{I}$ and $W_{O}$. In this paper, the attention outputs before $W_{AO}$ are used to preserve head-wise information, while the hidden states after $W_{O}$ are used, to represent a word. While it is common to use hidden states, in line with previous studies emphasizing the interpretability of attention weights, the outputs right after self-attention are explored as an alternative. This is because hidden states tend to lose such information through $W_{AO}$, $W_{I}$ and $W_{O}$.

Together with semantic sense, the study attempts to train mask for PoS or semantic role (SVO: Subject, Verb, Object), taking inspiration from \citet{Zhang2021} who tried to disentangle two distinct aspects by minimizing the overlap between masks using ${L}_{ovl}$ \footnote{To train distinct two masks $M^{(a)}$, $M^{(b)}$ for each layer $l$, $L_{ovl} = \frac{1}{|L|} \sum_{l \in L} \sum_{i,j} \mathbbm{1}(M_{i,j}^{(a)} + M_{i,j}^{(b)} > 1)$}. Unlike the prior study that masks a part of architecture and preserve the original embedding size, the masking approach in this work directly masks embeddings and reduces the dimensionality, akin to feature selection.

\subsubsection{Masking Self-attention Outputs}
\label{Masking Self-attention Outputs}

\begin{equation}
\begin{aligned}
{Sem}&{antic}_{attn}(x)\\
&= {Mask}_{attn}^{semantic}(k) \odot {attn}_{layer-wise}(x)
\end{aligned}
\end{equation}

\noindent A binary mask ${Mask}_{attn}^{semantic}(k)$ selects $k/a$ ($a$: attention head size) \footnote{If $a=64$ and $k=256$, the mask selects 4 heads.} heads per layer to get ${Semantic}_{attn}(x)$, which is the disentangled semantic sense representation of $x$. The mask has the same size as ${attn}_{layer-wise}(x)\in\mathbb{R}^{h \times l}$ ($h$: hidden size, $l$: number of layers, e.g., $h$=768 and $l$=12 in BERT$_{\texttt{base}}$), and reduces the size of ${attn}_{layer-wise}(x)$ from $\mathbb{R}^{h \times l}$ to $\mathbb{R}^{k \times l}$. 

\subsubsection{Masking Hidden States}
\label{Masking Hidden States}

\begin{equation}
\begin{aligned}
{Sem}&{antic}_{hs}(x)\\
&= {Mask}_{hs}^{semantic}(k) \odot {hs}_{layer-wise}(x)
\end{aligned}
\end{equation}

\noindent About hidden states, ${Mask}_{hs}^{semantic}(k)$ chooses $k$ dimensions from ${hs}_{layer-wise}(x)$ per layer to get ${Semantic}_{hs}(x)$.

\subsubsection{Loss}
\label{Loss}
The loss builds upon the \textit{Triplet Loss} framework proposed by \citet{Zhang2021}, who aimed to disentangle two aspects $a$ and $b$ from a text by training binary masks utilizing a \textit{straight-through} estimator \cite{DBLP:journals/corr/BengioLC13} over the architecture of PLM. It defines triplets $(x_0, x_1, x_2)$ that satisfy two conditions: (1) $y_0^{(a)} = y_1^{(a)} \not = y_2^{(a)}$ \footnote{It means $x_0$ and $x_1$ are same in terms of aspect $a$, while $x_0$ and $x_2$ are differ.}, and (2) $y_0^{(b)} \not = y_1^{(b)} = y_2^{(b)}$. The customized losses $L^{(a)}$ and $L^{(b)}$ are defined as follows. ($z^{(a)}$ and $z^{(b)}$ are the disentangled embeddings of $a$ and $b$ respectively.) 

\vspace{-0.5cm}
\begin{equation}
\begin{aligned}
&L^{(a)} = max(-cos(z_0^{(a)}, z_1^{(a)})+cos(z_0^{(a)}, z_2^{(a)}), 0)\\
&L^{(b)} = max(-cos(z_0^{(b)}, z_2^{(b)})+cos(z_0^{(b)}, z_1^{(b)}), 0)
\end{aligned} \footnote{The function $cos(x,y)$ computes the cosine similarity between two vectors $x$ and $y$.}
\end{equation}

\noindent The final loss depends on the number of aspects being disentangled, $L_{final} = L^{(a)}$ for a single aspect and $L_{final} = \frac{1}{2}\lambda(L^{(a)}+L^{(b)}) + (1-\lambda)L_{ovl}$  for two aspects. Here the hyperparameter $\lambda$ serves as a weighting factor, balancing the loss with $L_{ovl}$.

\subsection{Layer-wise Similarity Calculator}
\label{Layer-wise similarity calculator}
Given two sentences $s_1$ and $s_2$ containing a target word $w$, the similarity calculator $sim_{x,y}=f(x,y)$ takes representations of $w_{s_1}$ and $w_{s_2}$ ($w_{s_i}$ is $w$ in $s_i$) as inputs. It computes cosine similarity layer-wise, resulting in a vector $\in\mathbb{R}^{1 \times l}$ ($l$=12 in BERT$_{\texttt{base}}$).

\section{Experiment}
\subsection{Dataset}
This study uses three datasets: WiC \cite{pilehvar-camacho-collados-2019-wic}, CoarseWSD-20 \cite{loureiro2021analysis}, and SemCor \cite{miller-etal-1994-using}. WiC is formatted as binary classification without semantic sense labels (Figure \ref{fig:WiC example}), with train/dev/test splits and unlabeled test data. CoarseWSD-20 consists of 20 words with an average of 2.65 distinct semantic senses per word, with train/test splits of 23,370 and 10,196 sentences, respectively. SemCor consists of documents where some words in certain sentences are tagged with semantic senses. It includes a total of 276,878 sense-tagged sentences, 20,398 unique words, and an average of 1.64 distinct senses per word.

\vspace{-0.2cm}
\setlength{\tabcolsep}{5pt}
\renewcommand{\arraystretch}{1.2}
\begin{table}[htb!]
\resizebox{\columnwidth}{!}{%
{\small\begin{tabular}{lll|rrr}
\toprule
                                             &                                                  &            & \multicolumn{1}{c}{train} & \multicolumn{1}{c}{dev} & \multicolumn{1}{c}{test} \\ \hline

\multicolumn{2}{l|}{\multirow{3}{*}{Binary Classifer}}                                          & WiC          & 4,885                   & 543            & 638                    \\ \cline{3-3}
\multicolumn{2}{l|}{}                                                                           & CoarseWSD-20 & 180,000                & 20,000         & 100,000                \\ \cline{3-3}
\multicolumn{2}{l|}{}                                                                           & SemCor       & 285,088                 & 31,672         & 89,760                 \\ \hline
\multicolumn{1}{l|}{\multirow{6}{*}{Masker}} & \multicolumn{1}{l|}{\multirow{2}{*}{semantic}}   & CoarseWSD-20 & 168,264                & 18,696         & -                    \\ \cline{3-3}
\multicolumn{1}{l|}{}                        & \multicolumn{1}{l|}{}                            & SemCor       & 402,480                 & 44,712         & -                    \\ \cline{2-6} 
\multicolumn{1}{l|}{}                        & \multicolumn{1}{l|}{\multirow{2}{*}{(+) PoS}}    & CoarseWSD-20 & 163,320                & 18,144         & -                    \\ \cline{3-3}
\multicolumn{1}{l|}{}                        & \multicolumn{1}{l|}{}                            & SemCor       & 235,368                 & 26,144         & -                    \\ \cline{2-6} 
\multicolumn{1}{l|}{}                        & \multicolumn{1}{l|}{\multirow{2}{*}{(+) SVO}}    & CoarseWSD-20 & 16,704                 & 1,856          & -                    \\ \cline{3-3}
\multicolumn{1}{l|}{}                        & \multicolumn{1}{l|}{}                            & SemCor               & 38,768                  & 4,304                 & -                    \\ 
\bottomrule
\end{tabular}%
}}
\vspace{-0.3cm}
\caption{Processed data statistics of each dataset.}
\vspace{-0.3cm}
\label{tab:processed data}
\end{table}

\noindent For experimentation, three datasets are processed and train/dev/test subsets are extracted (Table \ref{tab:processed data}). See Section \ref{Data Processing} for the data processing method. 

\begin{figure*}[htb!]
\centering
\includegraphics[width=\textwidth]{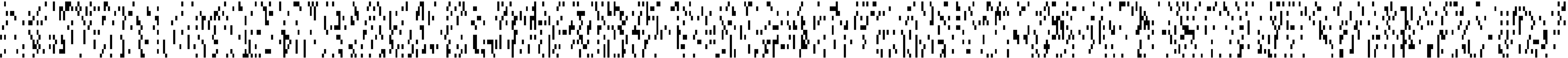}
\vspace{-0.6cm}
\caption{The visualization of $Mask_{hs}^{Semantic}(128) \in \mathbb{R}^{768 \times 12}$ which is the trained binary mask. Black denotes the selected(=1) dimensions, and white is unselected(=0) dimensions.}
\label{fig:mask_1_hs_128_semantic}
\vspace{-0.4cm}
\end{figure*}

\subsection{Experimental Details}
The masker and binary classifier are trained using a batch size of 8, Adam optimizer \cite{kingma2017adam}, and a learning rate of 0.01. Early stopping is utilized based on the dev set performance. The masker training explores different hidden states dimensions ($n\_dim=[128, 384, 512]$) and the number of attention heads ($n\_head=[4, 6, 8]$). The binary classifier consists of a linear layer followed by a sigmoid function. It takes cosine similarities between a pair and outputs a Boolean value indicating whether $w_{s_1}$ and $w_{s_2}$ share the same meaning. 

\subsection{Result}

\vspace{-0.1cm}
\setlength{\tabcolsep}{5pt}
\renewcommand{\arraystretch}{1.2}
\begin{table}[htb!]
\resizebox{\columnwidth}{!}{%
{\small\begin{tabular}{lll|rllrll}
\toprule

\multicolumn{3}{c|}{\textbf{WiC}}                  & \multicolumn{3}{c}{attention} & \multicolumn{3}{c}{hidden}    \\ \hline
\multicolumn{3}{l|}{Baseline (Cased BERT$_\texttt{base}$)}  & \multicolumn{3}{r}{0.682 \scriptsize{($\pm$ 0.002)}} & \multicolumn{3}{r}{\textcolor{blue}{0.651} \scriptsize{($\pm$ 0.003)}} \\ \cline{1-3}
\multicolumn{3}{l|}{(+) Layer-wise Calculator}     & \multicolumn{3}{r}{\textbf{0.705} \scriptsize{($\pm$ 0.002)}} & \multicolumn{3}{r}{0.667 \scriptsize{($\pm$ 0.002)}} \\  

\midrule \midrule

\multicolumn{3}{c|}{\textbf{CoarseWSD-20}} & \multicolumn{3}{c}{attention} & \multicolumn{3}{c}{hidden}    \\ \hline
\multicolumn{3}{l|}{Baseline (Cased BERT$_\texttt{base}$)}                                      & \multicolumn{3}{r}{0.754 \scriptsize{($\pm$ 0.001)}} & \multicolumn{3}{r}{\textcolor{blue}{0.743} \scriptsize{($\pm$ 0.000)}} \\ \cline{1-3}
\multicolumn{3}{l|}{(+) Layer-wise Calculator}                                            & \multicolumn{3}{r}{0.781 \scriptsize{($\pm$ 0.000)}} & \multicolumn{3}{r}{0.768 \scriptsize{($\pm$ 0.001)}} \\ \cline{1-3}
\multicolumn{1}{l|}{\multirow{3}{*}{\begin{tabular}[c]{@{}c@{}}(+) Disentanglement \\ by Masking\end{tabular}}} & \multicolumn{2}{l|}{semantic} & \multicolumn{3}{r}{0.788 \scriptsize{($\pm$ 0.007)}} & \multicolumn{3}{r}{\textbf{0.802}	\scriptsize{($\pm$ 0.002)}} \\ \cline{2-3}
\multicolumn{1}{l|}{}                                   & \multicolumn{2}{l|}{w/ PoS}    & \multicolumn{3}{r}{0.790	\scriptsize{($\pm$ 0.003)}} & \multicolumn{3}{r}{\textbf{0.802}	\scriptsize{($\pm$ 0.004)}} \\ \cline{2-3}
\multicolumn{1}{l|}{}                                   & \multicolumn{2}{l|}{w/ SVO}    & \multicolumn{3}{r}{0.796	\scriptsize{($\pm$ 0.006)}} & \multicolumn{3}{r}{0.793	\scriptsize{($\pm$ 0.006)}} \\ 

\midrule \midrule

\multicolumn{3}{c|}{\textbf{SemCor}} & \multicolumn{3}{c}{attention} & \multicolumn{3}{c}{hidden}    \\ \hline
\multicolumn{3}{l|}{Baseline (Cased BERT$_\texttt{base}$)}                                      & \multicolumn{3}{r}{0.777 \scriptsize{($\pm$ 0.001)}} & \multicolumn{3}{r}{\textcolor{blue}{0.758} \scriptsize{($\pm$ 0.007)}} \\ \cline{1-3}
\multicolumn{3}{l|}{(+) Layer-wise Calculator}                                            & \multicolumn{3}{r}{0.780 \scriptsize{($\pm$ 0.001)}} & \multicolumn{3}{r}{0.764 \scriptsize{($\pm$ 0.003)}} \\ \cline{1-3}
\multicolumn{1}{l|}{\multirow{3}{*}{\begin{tabular}[c]{@{}c@{}}(+) Disentanglement \\ by Masking\end{tabular}}} & \multicolumn{2}{l|}{semantic} & \multicolumn{3}{r}{0.774 \scriptsize{($\pm$ 0.001)}} & \multicolumn{3}{r}{\textbf{0.781}	\scriptsize{($\pm$ 0.001)}} \\ \cline{2-3}
\multicolumn{1}{l|}{}                                   & \multicolumn{2}{l|}{w/ PoS}    & \multicolumn{3}{r}{0.781	\scriptsize{($\pm$ 0.003)}} & \multicolumn{3}{r}{0.776	\scriptsize{($\pm$ 0.008)}} \\ \cline{2-3}
\multicolumn{1}{l|}{}                                   & \multicolumn{2}{l|}{w/ SVO}    & \multicolumn{3}{r}{0.775	\scriptsize{($\pm$ 0.004)}} & \multicolumn{3}{r}{0.776	\scriptsize{($\pm$ 0.005)}} \\ 
\bottomrule
\end{tabular}%
}}
\vspace{-0.2cm}
\caption{Mean accuracy and standard deviation in binary classification across three test sets for each dataset. Masker configuration: $n\_dim = 128$, $n\_head = 6$}
\vspace{-0.5cm}
\label{tab:performance}
\end{table}

\noindent Table \ref{tab:performance} compares the performance of the proposed approaches with the Baseline (highlighted in \textcolor{blue}{blue}), which sums the last four layers' hidden states to represent a word. For Figure \ref{fig:masker}, the Baseline uses the last four rows of \textcolor{red}{[1]}(attention) or \textcolor{red}{[3]}(hidden), the (+) Layer-wise Calculator uses the entire \textcolor{red}{[1]} or \textcolor{red}{[3]}, and the (+) Disentanglement by Masking approach utilizes \textcolor{red}{[2]} or \textcolor{red}{[4]}. WiC lacks semantic sense labels, so the evaluation is limited to the layer-wise calculator as it cannot be converted into triplets for masker training. The best score per dataset is in \textbf{bold}. Regardless of the dataset, layer-wise information improves performance by 1-2\% compared to the baseline. Moreover, incorporating disentangled embeddings further enhances hidden states (hidden) performance by around 2\%.

\begin{figure}[htb!]
\centering
\includegraphics[width=\columnwidth]{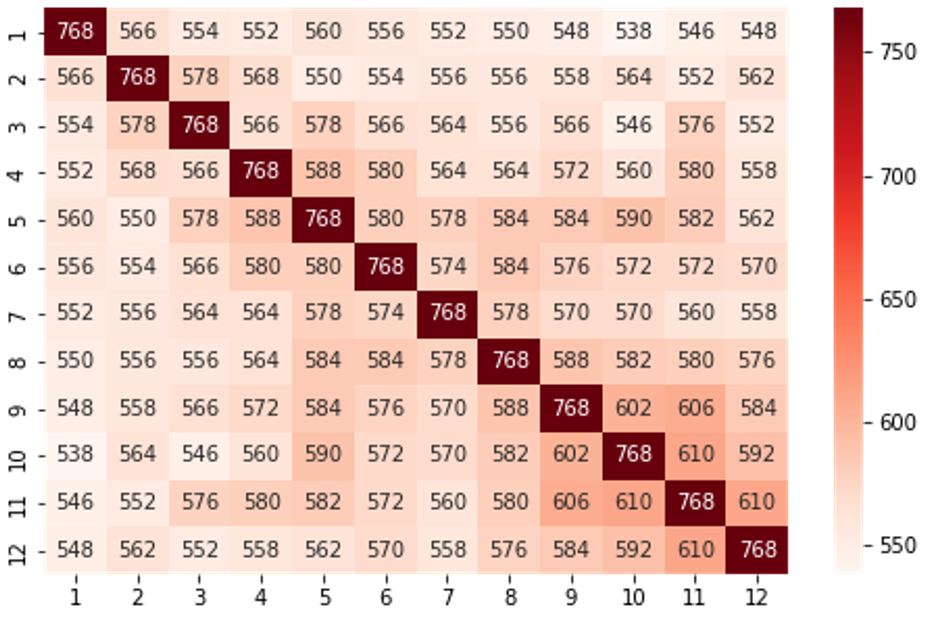}
\vspace{-0.8cm}
\caption{Number of equivalent values out of 768 dimensions across layers of trained mask depicted in Figure \ref{fig:mask_1_hs_128_semantic}}
\label{fig:mask_layer_sim_1_hs_128_semantic_2}
\vspace{-0.8cm}
\end{figure}

\begin{figure}[htb!]
\centering
\includegraphics[width=\columnwidth]{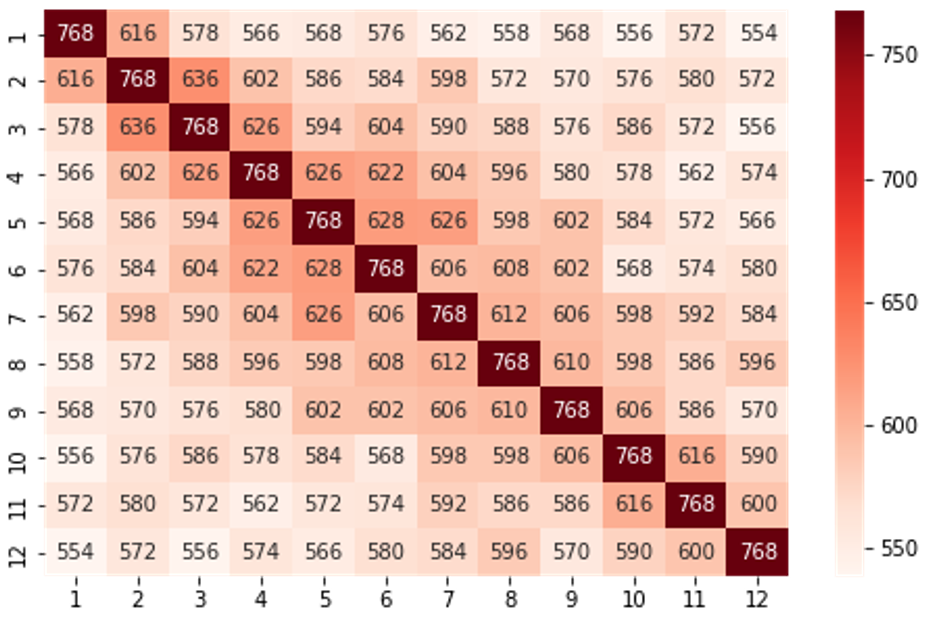}
\vspace{-0.7cm}
\caption{Number of dimensions with equivalent values across layers in $Mask_{hs}^{PoS}(128)$}
\label{fig:mask_layer_sim_2_hs_128_pos_2}
\vspace{-0.7cm}
\end{figure}

Despite being uncommon as word representations, self-attention outputs that preserve head-wise information seem to be inherently superior to hidden states for semantic representation. However, the masking approach proves more effective for hidden states, and disentangled hidden states perform the best among the experimental units. It is notable that there is improvement even with a 1/6 reduction in the embedding size from 768 to 128. In addition, adjacent layers tend to exhibit similar dimensions selected by the mask, particularly in the later layers for semantic sense (Figure \ref{fig:mask_layer_sim_1_hs_128_semantic_2}) and in the earlier layers for PoS (Figure \ref{fig:mask_layer_sim_2_hs_128_pos_2}), regardless of each dataset or implementation. While not generating the identical masks every time, the trained masks for hidden states shared these similar attributes. However, no such tendency is observed with attention outputs.

Using layer-wise information instead of collapsing it led to improved performance. The trained weights of binary classifier, which utilizes the mask shown in Figure \ref{fig:mask_1_hs_128_semantic}, were: [$-$2.914, $-$0.724, 1.403, 2.640, 3.660, 2.565, $-$3.712, $-$1.123, 0.081, 1.649, 0.518, 5.169]. Considering that the binary classifier determines the priority of information from each layer, it is observed that similarities between disentangled embeddings of layers 1, 2 and 7, 8 are rejected, while the last layer's similarity is most heavily reflected. Training PoS or SVO masks alongside the semantic sense mask, in anticipation of achieving additional improvement, did not yield the expected results. On the other hand, layer-wise information proves its own effectiveness, as our methodology achieved a 70.5\% accuracy on WiC testset, comparable to SenseBERT$_\texttt{BASE}$ \cite{levine-etal-2020-sensebert}, a model explicitly fine-tuned for sense disambiguation, which achieved a score of 70.3\%.

\vspace{-0.2cm}

\section{Conclusion}
The study investigated extracting the intrinsic semantic sense of words from pre-trained BERT$_{\texttt{base}}$ without modifying it, and uncovered the presence of noisy dimensions within embeddings to represent semantic sense. The merit of the masking approach in this paper can be extended to the ability to obtain aspect-specific representations by applying diverse learned masks to pre-existing offline saved embeddings, such as those utilized in a search engine system where embeddings per document are already indexed. Limitations and future directions are discussed in Section \ref{Limitation}.

\section{Limitation}
\label{Limitation}
The main limitation of this study is its focus on Cased BERT$_{\texttt{base}}$ exclusively. In experiments using Uncased BERT$_{\texttt{base}}$, the Baseline outperformed Cased BERT$_{\texttt{base}}$, which may explain why Uncased BERT$_{\texttt{base}}$ is commonly preferred as the backbone model for embedding-based WSD. However, the performance improvement of Uncased BERT$_{\texttt{base}}$ through disentanglement was marginal. 

The difference between Cased and Uncased models lies in how the tokenizer constructs its own vocabulary, specifically regarding whether or not the casing of tokens is preserved. The vocabulary of Cased PLM can be considered more fragmented (less normalized) compared to that of Uncased PLM. Cased tokens likely possess specific traits, such as their frequent usage for proper nouns like names of places or individuals. As the tokenizer logic itself influences word representation, contextual embeddings derived from cased tokens may contain unique information absent in those generated from uncased tokens. This hypothesis could be expanded to propose that there might be additional disentangled aspects within the embeddings. Nevertheless, this hypothesis remains speculative, and further research is required to investigate the relationship between vocabulary composition, differences in embeddings, and disentanglement methodologies.

\bibliography{acl_latex}

\begin{thebibliography}{26}
\expandafter\ifx\csname natexlab\endcsname\relax\def\natexlab#1{#1}\fi

\bibitem[{Bengio et~al.(2013)Bengio, L{\'{e}}onard, and Courville}]{DBLP:journals/corr/BengioLC13}
Yoshua Bengio, Nicholas L{\'{e}}onard, and Aaron~C. Courville. 2013.
\newblock \href {http://arxiv.org/abs/1308.3432} {Estimating or propagating gradients through stochastic neurons for conditional computation}.
\newblock \emph{CoRR}, abs/1308.3432.

\bibitem[{Bommasani et~al.(2020)Bommasani, Davis, and Cardie}]{bommasani-etal-2020-interpreting}
Rishi Bommasani, Kelly Davis, and Claire Cardie. 2020.
\newblock \href {https://doi.org/10.18653/v1/2020.acl-main.431} {{I}nterpreting {P}retrained {C}ontextualized {R}epresentations via {R}eductions to {S}tatic {E}mbeddings}.
\newblock In \emph{Proceedings of the 58th Annual Meeting of the Association for Computational Linguistics}, pages 4758--4781, Online. Association for Computational Linguistics.

\bibitem[{Carbonneau et~al.(2022{\natexlab{a}})Carbonneau, Zaïdi, Boilard, and Gagnon}]{9947342}
Marc-André Carbonneau, Julian Zaïdi, Jonathan Boilard, and Ghyslain Gagnon. 2022{\natexlab{a}}.
\newblock \href {https://doi.org/10.1109/TNNLS.2022.3218982} {Measuring disentanglement: A review of metrics}.
\newblock \emph{IEEE Transactions on Neural Networks and Learning Systems}, pages 1--15.

\bibitem[{Carbonneau et~al.(2022{\natexlab{b}})Carbonneau, Zaïdi, Boilard, and Gagnon}]{Carbonneau}
Marc-André Carbonneau, Julian Zaïdi, Jonathan Boilard, and Ghyslain Gagnon. 2022{\natexlab{b}}.
\newblock \href {https://doi.org/10.1109/TNNLS.2022.3218982} {Measuring disentanglement: A review of metrics}.
\newblock \emph{IEEE Transactions on Neural Networks and Learning Systems}, pages 1--15.

\bibitem[{Chen et~al.(2016)Chen, Duan, Houthooft, Schulman, Sutskever, and Abbeel}]{chen2016infogan}
Xi~Chen, Yan Duan, Rein Houthooft, John Schulman, Ilya Sutskever, and Pieter Abbeel. 2016.
\newblock Infogan: Interpretable representation learning by information maximizing generative adversarial nets.
\newblock \emph{Advances in neural information processing systems}, 29.

\bibitem[{Clark et~al.(2019)Clark, Khandelwal, Levy, and Manning}]{clark2019does}
Kevin Clark, Urvashi Khandelwal, Omer Levy, and Christopher~D Manning. 2019.
\newblock What does bert look at? an analysis of bert's attention.
\newblock \emph{arXiv preprint arXiv:1906.04341}.

\bibitem[{Colombo et~al.(2021)Colombo, Clavel, and Piantanida}]{DBLP:journals/corr/abs-2105-02685}
Pierre Colombo, Chlo{\'{e}} Clavel, and Pablo Piantanida. 2021.
\newblock \href {http://arxiv.org/abs/2105.02685} {A novel estimator of mutual information for learning to disentangle textual representations}.
\newblock \emph{CoRR}, abs/2105.02685.

\bibitem[{Devlin et~al.(2019)Devlin, Chang, Lee, and Toutanova}]{devlin-etal-2019-bert}
Jacob Devlin, Ming-Wei Chang, Kenton Lee, and Kristina Toutanova. 2019.
\newblock \href {https://doi.org/10.18653/v1/N19-1423} {{BERT}: Pre-training of deep bidirectional transformers for language understanding}.
\newblock In \emph{Proceedings of the 2019 Conference of the North {A}merican Chapter of the Association for Computational Linguistics: Human Language Technologies, Volume 1 (Long and Short Papers)}, pages 4171--4186, Minneapolis, Minnesota. Association for Computational Linguistics.

\bibitem[{Do and Tran(2019)}]{do2019theory}
Kien Do and Truyen Tran. 2019.
\newblock Theory and evaluation metrics for learning disentangled representations.
\newblock \emph{arXiv preprint arXiv:1908.09961}.

\bibitem[{Hewitt and Manning(2019)}]{hewitt-manning-2019-structural}
John Hewitt and Christopher~D. Manning. 2019.
\newblock \href {https://doi.org/10.18653/v1/N19-1419} {{A} structural probe for finding syntax in word representations}.
\newblock In \emph{Proceedings of the 2019 Conference of the North {A}merican Chapter of the Association for Computational Linguistics: Human Language Technologies, Volume 1 (Long and Short Papers)}, pages 4129--4138, Minneapolis, Minnesota. Association for Computational Linguistics.

\bibitem[{Jawahar et~al.(2019)Jawahar, Sagot, and Seddah}]{jawahar-etal-2019-bert}
Ganesh Jawahar, Beno{\^\i}t Sagot, and Djam{\'e} Seddah. 2019.
\newblock \href {https://doi.org/10.18653/v1/P19-1356} {What does {BERT} learn about the structure of language?}
\newblock In \emph{Proceedings of the 57th Annual Meeting of the Association for Computational Linguistics}, pages 3651--3657, Florence, Italy. Association for Computational Linguistics.

\bibitem[{Kingma and Ba(2017)}]{kingma2017adam}
Diederik~P. Kingma and Jimmy Ba. 2017.
\newblock \href {http://arxiv.org/abs/1412.6980} {Adam: A method for stochastic optimization}.

\bibitem[{Levine et~al.(2020)Levine, Lenz, Dagan, Ram, Padnos, Sharir, Shalev-Shwartz, Shashua, and Shoham}]{levine-etal-2020-sensebert}
Yoav Levine, Barak Lenz, Or~Dagan, Ori Ram, Dan Padnos, Or~Sharir, Shai Shalev-Shwartz, Amnon Shashua, and Yoav Shoham. 2020.
\newblock \href {https://doi.org/10.18653/v1/2020.acl-main.423} {{S}ense{BERT}: Driving some sense into {BERT}}.
\newblock In \emph{Proceedings of the 58th Annual Meeting of the Association for Computational Linguistics}, pages 4656--4667, Online. Association for Computational Linguistics.

\bibitem[{Loureiro et~al.(2021)Loureiro, Rezaee, Pilehvar, and Camacho-Collados}]{loureiro2021analysis}
Daniel Loureiro, Kiamehr Rezaee, Mohammad~Taher Pilehvar, and Jose Camacho-Collados. 2021.
\newblock Analysis and evaluation of language models for word sense disambiguation.
\newblock \emph{Computational Linguistics}, 47(2):387--443.

\bibitem[{Miller et~al.(1994)Miller, Chodorow, Landes, Leacock, and Thomas}]{miller-etal-1994-using}
George~A. Miller, Martin Chodorow, Shari Landes, Claudia Leacock, and Robert~G. Thomas. 1994.
\newblock \href {https://aclanthology.org/H94-1046} {Using a semantic concordance for sense identification}.
\newblock In \emph{{H}uman {L}anguage {T}echnology: Proceedings of a Workshop held at {P}lainsboro, {N}ew {J}ersey, {M}arch 8-11, 1994}.

\bibitem[{Pilehvar and Camacho-Collados(2019)}]{pilehvar-camacho-collados-2019-wic}
Mohammad~Taher Pilehvar and Jose Camacho-Collados. 2019.
\newblock \href {https://doi.org/10.18653/v1/N19-1128} {{W}i{C}: the word-in-context dataset for evaluating context-sensitive meaning representations}.
\newblock In \emph{Proceedings of the 2019 Conference of the North {A}merican Chapter of the Association for Computational Linguistics: Human Language Technologies, Volume 1 (Long and Short Papers)}, pages 1267--1273, Minneapolis, Minnesota. Association for Computational Linguistics.

\bibitem[{Raganato et~al.(2017)Raganato, Camacho-Collados, and Navigli}]{raganato-etal-2017-word}
Alessandro Raganato, Jose Camacho-Collados, and Roberto Navigli. 2017.
\newblock \href {https://aclanthology.org/E17-1010} {Word sense disambiguation: A unified evaluation framework and empirical comparison}.
\newblock In \emph{Proceedings of the 15th Conference of the {E}uropean Chapter of the Association for Computational Linguistics: Volume 1, Long Papers}, pages 99--110, Valencia, Spain. Association for Computational Linguistics.

\bibitem[{Scarlini et~al.(2020{\natexlab{a}})Scarlini, Pasini, and Navigli}]{SensEmBERT}
Bianca Scarlini, Tommaso Pasini, and Roberto Navigli. 2020{\natexlab{a}}.
\newblock \href {https://doi.org/10.1609/aaai.v34i05.6402} {Sensembert: Context-enhanced sense embeddings for multilingual word sense disambiguation}.
\newblock volume~34.

\bibitem[{Scarlini et~al.(2020{\natexlab{b}})Scarlini, Pasini, and Navigli}]{scarlini-etal-2020-contexts}
Bianca Scarlini, Tommaso Pasini, and Roberto Navigli. 2020{\natexlab{b}}.
\newblock \href {https://doi.org/10.18653/v1/2020.emnlp-main.285} {With more contexts comes better performance: Contextualized sense embeddings for all-round word sense disambiguation}.
\newblock In \emph{Proceedings of the 2020 Conference on Empirical Methods in Natural Language Processing (EMNLP)}, pages 3528--3539, Online. Association for Computational Linguistics.

\bibitem[{Tenney et~al.(2019)Tenney, Das, and Pavlick}]{tenney-etal-2019-bert}
Ian Tenney, Dipanjan Das, and Ellie Pavlick. 2019.
\newblock \href {https://doi.org/10.18653/v1/P19-1452} {{BERT} rediscovers the classical {NLP} pipeline}.
\newblock In \emph{Proceedings of the 57th Annual Meeting of the Association for Computational Linguistics}, pages 4593--4601, Florence, Italy. Association for Computational Linguistics.

\bibitem[{Vig and Belinkov(2019)}]{vig-belinkov-2019-analyzing}
Jesse Vig and Yonatan Belinkov. 2019.
\newblock \href {https://doi.org/10.18653/v1/W19-4808} {Analyzing the structure of attention in a transformer language model}.
\newblock In \emph{Proceedings of the 2019 ACL Workshop BlackboxNLP: Analyzing and Interpreting Neural Networks for NLP}, pages 63--76, Florence, Italy. Association for Computational Linguistics.

\bibitem[{Vishnubhotla et~al.(2021)Vishnubhotla, Hirst, and Rudzicz}]{vishnubhotla2021evaluation}
Krishnapriya Vishnubhotla, Graeme Hirst, and Frank Rudzicz. 2021.
\newblock An evaluation of disentangled representation learning for texts.
\newblock In \emph{Findings of the Association for Computational Linguistics: ACL-IJCNLP 2021}, pages 1939--1951.

\bibitem[{Xu et~al.(2020)Xu, Cheung, and Cao}]{xu2020variational}
Peng Xu, Jackie Chi~Kit Cheung, and Yanshuai Cao. 2020.
\newblock On variational learning of controllable representations for text without supervision.
\newblock In \emph{International Conference on Machine Learning}, pages 10534--10543. PMLR.

\bibitem[{Zhang et~al.(2021)Zhang, Meent, and Wallace}]{Zhang2021}
Xiongyi Zhang, Jan-Willem Meent, and Byron Wallace. 2021.
\newblock \href {https://doi.org/10.18653/v1/2021.emnlp-main.60} {Disentangling representations of text by masking transformers}.
\newblock pages 778--791.

\bibitem[{Zhao et~al.(2020)Zhao, Lin, Mi, Jaggi, and Schütze}]{Zhao2020}
Mengjie Zhao, Tao Lin, Fei Mi, Martin Jaggi, and Hinrich Schütze. 2020.
\newblock \href {https://doi.org/10.18653/v1/2020.emnlp-main.174} {Masking as an efficient alternative to finetuning for pretrained language models}.
\newblock pages 2226--2241.

\bibitem[{Zhao and Bethard(2020)}]{zhao-bethard-2020-berts}
Yiyun Zhao and Steven Bethard. 2020.
\newblock \href {https://doi.org/10.18653/v1/2020.acl-main.429} {How does {BERT}{'}s attention change when you fine-tune? an analysis methodology and a case study in negation scope}.
\newblock In \emph{Proceedings of the 58th Annual Meeting of the Association for Computational Linguistics}, pages 4729--4747, Online. Association for Computational Linguistics.

\end{thebibliography}
\bibliographystyle{acl_natbib}

\appendix
\section{Appendix}
\label{sec:appendix}

\subsection{Data Processing}
\label{Data Processing}
Raw train data of WiC and CoarseWSD-20 is split into train/dev (9:1 ratio), using CoarseWSD-20 test data and WiC dev data for test respectively. SemCor is divided into train/test (8:2 ratio), with the train set further split into train/dev (9:1 ratio). For CoarseWSD-20 and SemCor, triplets are sampled to train masker (Section \ref{Loss}). PoS and SVO labels for words are tagged by Spacy \footnote{https://github.com/explosion/spaCy} and Textacy \footnote{https://github.com/chartbeat-labs/textacy}. Furthermore, True- and false-labeled pairs are sampled for binary classification, following the format of WiC. To ensure robust evaluation, three distinct processed datasets are prepared, each of equal size.

\end{document}